\documentclass{article}
\usepackage{ijcai23}

\usepackage{times}
\usepackage{soul}
\usepackage{url}
\usepackage[hidelinks]{hyperref}
\usepackage[utf8]{inputenc}
\usepackage[small]{caption}
\usepackage{graphicx}
\usepackage{amsmath}
\usepackage{amsthm}
\usepackage{booktabs}
\usepackage{algorithm}
\usepackage{algorithmic}
\usepackage[switch]{lineno}

\title{MuLMINet: Multi-Layer Multi-Input Transformer Network with Weighted Loss}

\author{
Minwoo Seong$^{1, \ddagger}$
\and
Jeongseok Oh$^{1, \ddagger}$\And
SeungJun Kim$^{1*}$
\affiliations
$^1$ Gwangju Institute of Science and Technology\\
\emails
\{seongminwoo, jungseok17\}@gm.gist.ac.kr,
seungjun@gist.ac.kr
\emails
$^{\ddagger}$ These authors contributed equally to this work.
$^{*}$ Correspondence Author
}

\begin{document}

\maketitle

\begin{abstract}
    The increasing use of artificial intelligence (AI) technology in turn-based sports, such as badminton, has sparked significant interest in evaluating strategies through the analysis of match video data. Predicting future shots based on past ones plays a vital role in coaching and strategic planning. In this study, we present a Multi-Layer Multi-Input Transformer Network (MuLMINet) that leverages professional badminton player match data to accurately predict future shot types and area coordinates. Our approach resulted in achieving the runner-up (2nd place) in the IJCAI CoachAI Badminton Challenge 2023, Track 2. To facilitate further research, we have made our code publicly accessible online, contributing to the broader research community's knowledge and advancements in the field of AI-assisted sports analysis.
\end{abstract}

\section{Introduction}
The advent of artificial intelligence is bringing significant changes to sports analytics, with machine learning and deep learning playing key roles in this change \cite{ShuttleSet22}\cite{azad2022programmatic}. These techniques are not only employed for evaluating the capabilities and performance of individual players \cite{hulsmann2018classification}\cite{oagaz2021performance}\cite{ghosh2022decoach}, but they also play a crucial role in analyzing and assessing the strategies that emerge from interactions between players \cite{azad2022programmatic}\cite{han2022logistic}\cite{chen2022reliable}\cite{chang2023will}\cite{ShuttleNet_AAAI_2022}\cite{wang2022stroke}.
There have been several attempts to utilize artificial intelligence (AI) technology to evaluate players' strategies, especially in turn-based sports such as badminton. In particular, the ability to predict future shots based on past ones is integral to player coaching and strategic planning \cite{wang2021exploring}. The analysis of shot types that yield successful returns and the identification of optimal ball drop locations can significantly aid players in decision-making processes, thereby enhancing their performance.
Several attempts have been made to utilize AI technology to evaluate players' strategies, especially in turn-based sports such as badminton. Wang et al. proposed a language known as Badminton Language from Shot to Rally (BLSR) to represent the rally process and employed Convolutional Neural Networks (CNN) and encoder-based networks to predict shot influence \cite{wang2021exploring} \cite{wang2022stroke}. Wang et al. also proposed a ShuttleNet network, which predicts the sequence of badminton matches based on a transformer model.
In previously proposed models, the prediction of future strokes was primarily based on two variables: shot type and area coordinates. However, these features may not fully capture the complexity of the game as there are numerous factors that influence shot type and area. Recognizing this limitation, we propose a new BLSR prediction network, Multi-Layer Multi-Input Transformer Network (MuLMINet). We introduced our approach in the context of the IJCAI-2023 CoachAI Badminton Challenge Track 2: Forecasting Future Turn-Based Strokes in Badminton Rallies and won second position. This paper delineates our proposed methodology and the corresponding evaluation process, and the code is available on our GitHub repository: https://github.com/stan5dard/IJCAI-CoachAI-Challenge-2023.

\section{Method}
\subsection{Evaluation Metrics of IJCAI CoachAI Badminton Challenge Track 2}
In this challenge, the objective is to predict the future stroke shot type and area coordinates based on the data from the first 4 strokes. The evaluation metrics used for this task are defined as follows:

\begin{equation}
    \resizebox{0.5\linewidth}{!}{$Score=min(l_1, l_2, ..., l_6)$}
\end{equation}

\begin{equation}
    \resizebox{0.5\linewidth}{!}{$l_i=AVG(CE + MAE)$}
\end{equation}

The shot type is evaluated using the Cross-Entropy (CE) loss, which measures the discrepancy between the predicted and actual shot-type probabilities. The area coordinate is evaluated using the Mean Absolute Error (MAE), which quantifies the average difference between the predicted and actual area coordinates.
To calculate the final score, we take the average of six evaluation metrics and select the minimum value as the final score. This score serves as an overall measure of the predictive model's performance in predicting stroke shot type and landing coordination. The team with the lowest loss emerges as the winner of this challenge.

\subsection{Dataset and Data Preprocessing}

The competition dataset was based on the ShuttleSet dataset \cite{ShuttleSet22} and consisted of three separate files: train.csv, validation.csv, and test.csv. The dataset included a total of 30,172 strokes in the training set, 1,400 strokes in the validation set, and 2,040 strokes in the test set.
The training dataset, which included data with a rally length of five or more, formed the basis for model development. In contrast, the validation and test datasets only included data for the first four rallies. Participants were required to construct their own models using the training dataset, and these models were then tested on the validation and test sets as part of the evaluation process. The dataset is divided into four primary categories.

\begin{itemize}
    \item Rally category: round score A, round score B, get point player, and lose reason
    \item Temporal category: time number and frame number
    \item Spatial category: player location area, player location x, player location y, opponent location area, opponent location x, opponent location y, landing area, landing x, landing y, and landing height.
    \item Hitting category: backhand, around head, and shot type
\end{itemize}

In order to determine the input features for the model, we examined the correlation between variables. For categorical variables, we calculated the Cramer's V correlation matrix. Cramer's V is a measure of association used to assess the strength and direction of the relationship between two categorical variables.

\begin{figure}[htb!]
\centering
\includegraphics[width=\linewidth]{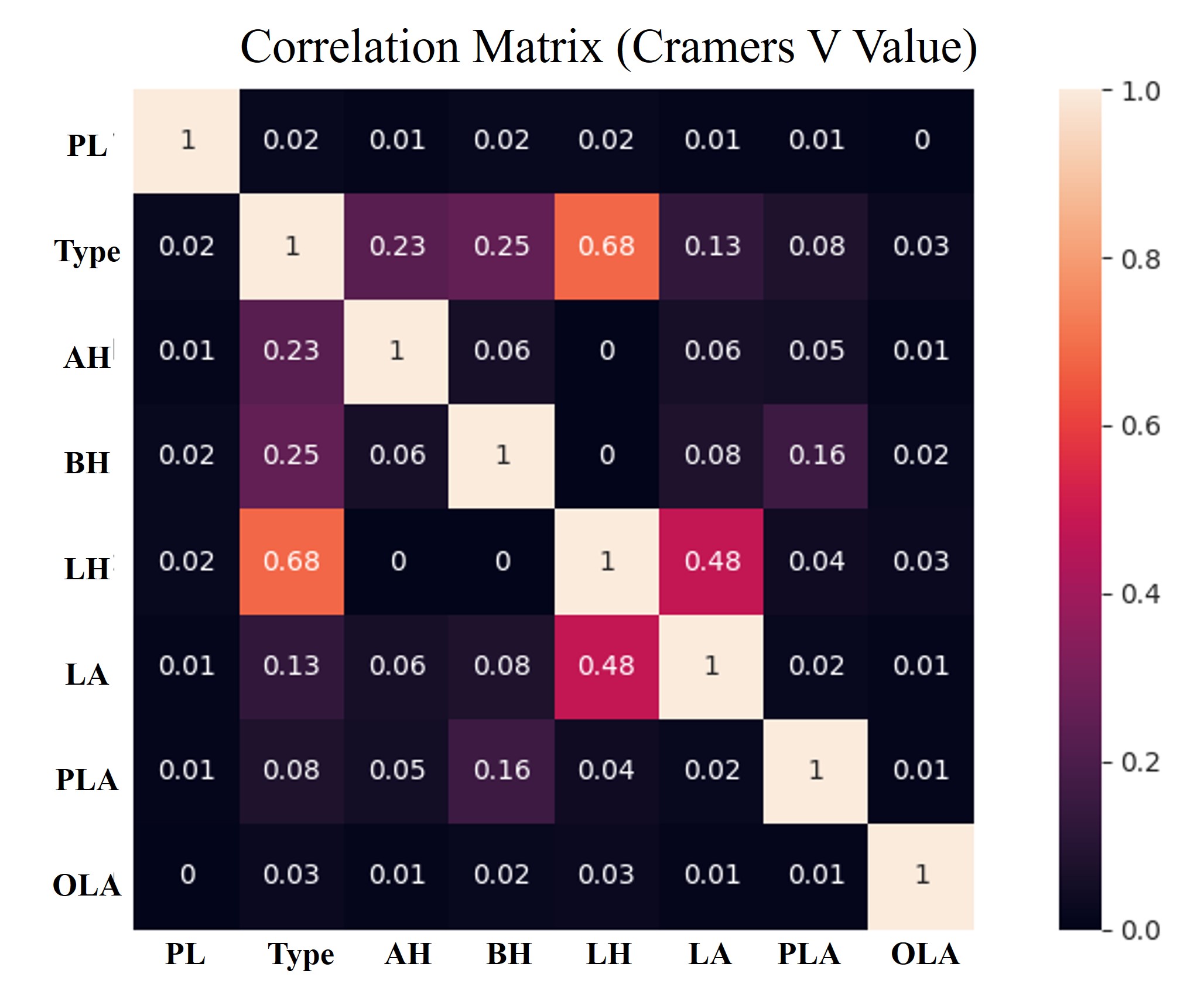}
\caption{Cramer's V Correlation Matrix between features: PL denotes player, AH denotes aroundhead, BH denotes backhand, LH denotes landing height, LA denotes landing area, PLA denotes player location area, and OLA denotes opponent location area}
\label{fig:Corr}
\end{figure}

Variables such as "aroundhead" (Cramér's V = 0.23) and "backhand" (Cramér's V = 0.25) demonstrated moderate correlations with shot type. On the other hand, landing height exhibited a strong correlation (Cramér's V = 0.68) with shot type. Based on these findings, we identified "aroundhead," "backhand," and landing height as additional input features for our model.

Moreover, for the prediction of area coordinates, we included player area location and opponent area location as additional features. As a result, we utilized a total of eight features as input variables for our predictive model. This selection of features was guided by the observed correlations and their potential relevance in predicting the desired outcomes. By incorporating these variables, we aimed to enhance the model's accuracy and predictive capabilities.

\begin{figure*}[htb!]
\centering
\includegraphics[width=\linewidth]{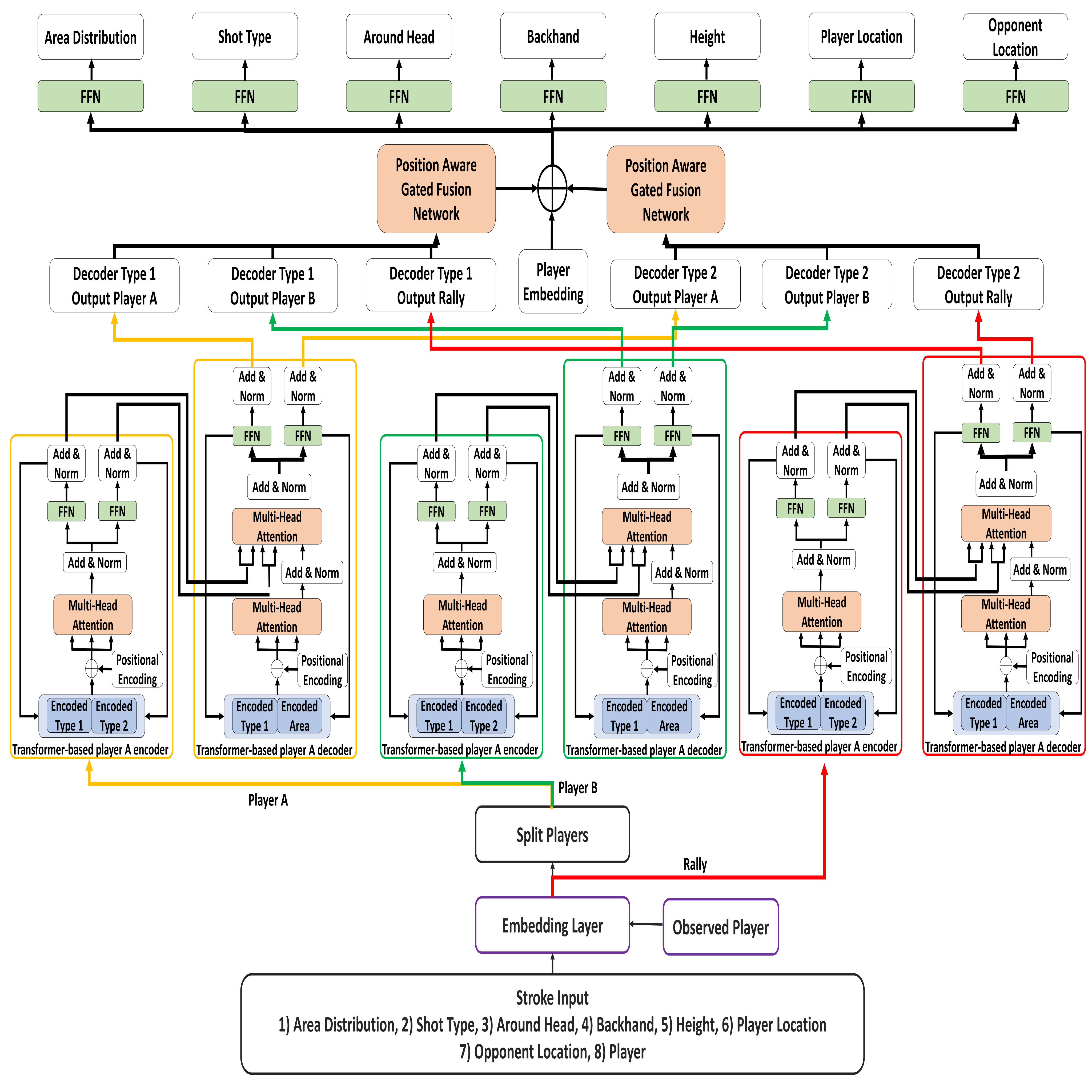}
\caption{Network Architecture: MuLMINet}
\label{fig:Network}
\end{figure*}

\subsection{Network Architecture Design}

The network architecture was specifically designed to accommodate additional input features (see Figure \ref{fig:Network}). We built our network based on the principles established by ShuttleNet, which incorporated player and rally encoders and decoders using BLSR data. We primarily used two encoding methods to process the input data.

Encoded Type 1 involved representing the shot type, aroundhead, backhand, height, player location, opponent location, and player variables within the input vector. By incorporating these variables together, we aimed to capture the relevant characteristics of the shot type and its associated attributes. Encoded Type 2 focused on encoding the area coordinate, aroundhead, backhand, height, player location, opponent location, and player variables. By including these features within the input vector, we aimed to capture the spatial information and positional attributes related to the area coordinates.

This design choice was motivated by the observed correlations between shot type, area coordinate, and other variables. By embedding these variables together, we aimed to achieve improved prediction performance. Similarly, for the decoder output, Type 1 and Type 2 were each passed through the Position Aware Gated Fusion Network. This network was responsible for predicting area, shot type, aroundhead, backhand, height, player location, and opponent location for each observed player.

\subsection{Loss Function Design and Hyper-parameter tuning}

We designed our total loss function as follows:
\begin{equation}
    \resizebox{0.9\linewidth}{!}{$L_{T}$=$\alpha$($L_{ST}$+$L_{SL}$) + (1-$\alpha$)($L_{B}$+$L_{A}$+$L_{H}$+$L_{PL}$+$L_{OL}$)}
\end{equation}

$L_{T}$ denotes total loss, $L_{ST}$ denotes loss of shot type, $L_{SL}$ denotes loss of shot landing coordination, $L_{B}$ denotes loss of backhand, $L_{A}$ denotes loss of around-head, $L_{H}$ denotes loss of landing height, $L_{PL}$ denotes loss of player location area, and $L_{OL}$ denotes loss of opponent location area. Alpha ($\alpha$) is a hyperparameter that controls the trade-off between the shot type/location loss and the other losses.

The model was evaluated using a dataset divided into three subsets: train set, validation set, and test set, derived from the train.csv dataset. The test set remained fixed, while the train set and validation set were evaluated using a 5-fold cross-validation approach. During the evaluation process, the model underwent 300 epochs with a fixed learning rate of 0.0001 and a batch size of 32. Hyper-parameter evaluation was conducted by varying the Dimension and Alpha values (see Table \ref{tab:Hyper-parameter}). Scores were computed based on the evaluation metric specified in the challenge, and the results were summarized in a table. The average value and standard deviation of the scores were calculated to assess the model's performance.

\begin{table}[ht]
\centering
{
\begin{tabular}{ll}
\toprule
Hyper-parameter & Values \\
\midrule
Learning rate & 0.0001 \\
Batch size & 32 \\
Dimension & 32, 64, 128 \\
Alpha & 0.3, 0.35, 0.4, 0.45 \\
Layer number & 1, 2, 3 \\
Epoch & 300 \\

\bottomrule
\end{tabular}
}
\caption{\label{tab:Hyper-parameter} Hyper-parameter for MuLMINet evaluation}
\end{table} 

\section{Results}
\subsection{Loss Selection Module}

\begin{figure}[htb!]
\centering
\includegraphics[width=\linewidth]{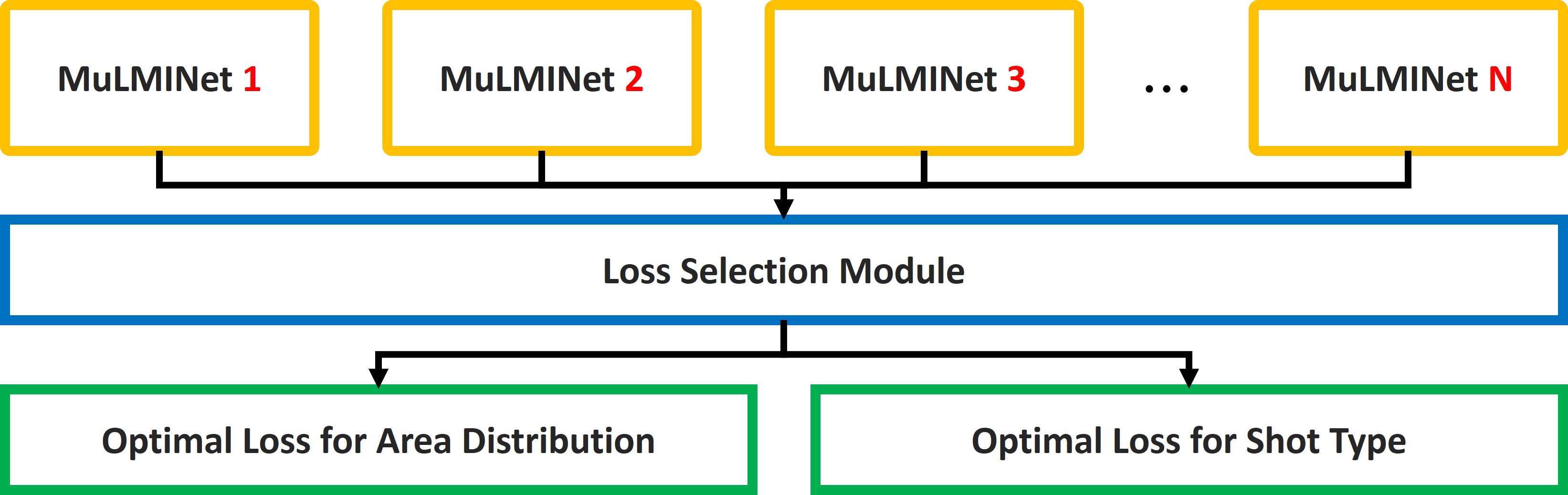}
\caption{Loss Selection Module}
\label{fig:LossSelection}
\end{figure}

To identify the optimal classifier and regressor, we implemented a Loss Selection Module (see Figure \ref{fig:LossSelection}) that evaluated 72 different hyperparameter cases. The purpose of this module was to determine the ideal combination of models by comparing the average loss across 5-fold cross-validation. For each hyperparameter combination, we assessed the average loss and identified the combinations that exhibited accurate predictions for shot type and area coordinates. These selected combinations were then submitted to the final challenge as our model entries, reflecting the comprehensive evaluation conducted through the Loss Selection Module.

\subsection{Optimal Parameter for MuLMINet}

After implementing the Loss Selection Module, we successfully identified the best case through a comprehensive evaluation of 36 different hyperparameter combinations. Based on the evaluation results, we selected the combinations that exhibited accurate predictions for shot type and area coordinates. The final combination was submitted as a model entry for the challenge, and Table \ref{tab:Optimal-parameter} shows the score from the validation process and the final score of the challenge.

\begin{table}[ht]
\centering
{
\begin{tabular}{llll}
\toprule

Phase & Total loss & Area loss & Shot loss \\ 
\midrule

Validation phase & 2.489 & 0.6674 & 1.8216 \\
Testing phase & 2.5830 & 0.7703 & 1.8127 \\
		
\bottomrule
\end{tabular}
}
\caption{\label{tab:Optimal-parameter} Loss score of validation and testing phase}
\end{table} 

\section{Discussions}

In the context of our study, it is worth exploring the potential benefits of treating the inputs for area and shot type differently based on their correlations. Although our initial approach involved embedding all features together in Encoded Type 1 and Encoded Type 2, regardless of their associations with shot type and area, we acknowledge that incorporating highly relevant features for each prediction task could yield improvements in accuracy.

For future research, we recommend exploring alternative embedding strategies that take into account feature correlations and tailoring them to the specific requirements of each prediction task. Such investigations could provide insights into optimizing model performance and achieving even lower loss in area coordinates and shot-type prediction tasks.

\section{Conclusions}

In conclusion, we propose MuLMINet, a novel architecture that leverages multi-layer and multi-input approaches to accurately predict both shot types and areas on the BSLR dataset. By integrating a weighted sum loss function and a loss selection module, we derive optimal parameters and models. As a result, we achieve the runner-up position (2nd winner) in the IJCAI CoachAI Badminton Challenge.

\appendix

\section*{Acknowledgments}
This work was supported by the GIST-MIT Research Collaboration grant, which was funded by GIST in 2023. We appreciate the high-performance GPU computing support of HPC-AI Open Infrastructure via GIST SCENT. 

\bibliographystyle{named}
\bibliography{ijcai23}

\begin{thebibliography}{}

\bibitem[\protect\citeauthoryear{Azad \bgroup \em et al.\egroup }{2022}]{azad2022programmatic}
Abdus~Salam Azad, Edward Kim, Qiancheng Wu, Kimin Lee, Ion Stoica, Pieter Abbeel, Alberto Sangiovanni-Vincentelli, and Sanjit~A Seshia.
\newblock Programmatic modeling and generation of real-time strategic soccer environments for reinforcement learning.
\newblock In {\em Proceedings of the AAAI Conference on Artificial Intelligence}, volume~36, pages 6028--6036, 2022.

\bibitem[\protect\citeauthoryear{Chang \bgroup \em et al.\egroup }{2023}]{chang2023will}
Kai-Shiang Chang, Wei-Yao Wang, and Wen-Chih Peng.
\newblock Where will players move next? dynamic graphs and hierarchical fusion for movement forecasting in badminton.
\newblock In {\em Proceedings of the AAAI Conference on Artificial Intelligence}, volume~37, pages 6998--7005, 2023.

\bibitem[\protect\citeauthoryear{Chen \bgroup \em et al.\egroup }{2022}]{chen2022reliable}
Xiusi Chen, Jyun-Yu Jiang, Kun Jin, Yichao Zhou, Mingyan Liu, P~Jeffrey Brantingham, and Wei Wang.
\newblock Reliable: Offline reinforcement learning for tactical strategies in professional basketball games.
\newblock In {\em Proceedings of the 31st ACM International Conference on Information \& Knowledge Management}, pages 3023--3032, 2022.

\bibitem[\protect\citeauthoryear{Du and Peng}{2023}]{ShuttleSet22}
Wei{-}Yao Wang and~Wei{-}Wei Du and Wen{-}Chih Peng.
\newblock Shuttleset22: Benchmarking stroke forecasting with stroke-level badminton dataset.
\newblock {\em CoRR}, abs/2306.15664, 2023.

\bibitem[\protect\citeauthoryear{Ghosh \bgroup \em et al.\egroup }{2022}]{ghosh2022decoach}
Indrajeet Ghosh, Sreenivasan~Ramasamy Ramamurthy, Avijoy Chakma, and Nirmalya Roy.
\newblock Decoach: Deep learning-based coaching for badminton player assessment.
\newblock {\em Pervasive and Mobile Computing}, 83:101608, 2022.

\bibitem[\protect\citeauthoryear{Han \bgroup \em et al.\egroup }{2022}]{han2022logistic}
Yewon Han, Jaeho Kim, Hon Keung~Tony Ng, and Seong~W Kim.
\newblock Logistic regression model for a bivariate binomial distribution with applications in baseball data analysis.
\newblock {\em Entropy}, 24(8):1138, 2022.

\bibitem[\protect\citeauthoryear{H{\"u}lsmann \bgroup \em et al.\egroup }{2018}]{hulsmann2018classification}
Felix H{\"u}lsmann, Jan~Philip G{\"o}pfert, Barbara Hammer, Stefan Kopp, and Mario Botsch.
\newblock Classification of motor errors to provide real-time feedback for sports coaching in virtual reality—a case study in squats and tai chi pushes.
\newblock {\em Computers \& Graphics}, 76:47--59, 2018.

\bibitem[\protect\citeauthoryear{Oagaz \bgroup \em et al.\egroup }{2021}]{oagaz2021performance}
Hawkar Oagaz, Breawn Schoun, and Min-Hyung Choi.
\newblock Performance improvement and skill transfer in table tennis through training in virtual reality.
\newblock {\em IEEE Transactions on Visualization and Computer Graphics}, 28(12):4332--4343, 2021.

\bibitem[\protect\citeauthoryear{Wang \bgroup \em et al.\egroup }{2021}]{wang2021exploring}
Wei-Yao Wang, Teng-Fong Chan, Hui-Kuo Yang, Chih-Chuan Wang, Yao-Chung Fan, and Wen-Chih Peng.
\newblock Exploring the long short-term dependencies to infer shot influence in badminton matches.
\newblock In {\em 2021 IEEE International Conference on Data Mining (ICDM)}, pages 1397--1402. IEEE, 2021.

\bibitem[\protect\citeauthoryear{Wang \bgroup \em et al.\egroup }{2022a}]{wang2022stroke}
Wei-Yao Wang, Teng-Fong Chan, Wen-Chih Peng, Hui-Kuo Yang, Chih-Chuan Wang, and Yao-Chung Fan.
\newblock How is the stroke? inferring shot influence in badminton matches via long short-term dependencies.
\newblock {\em ACM Transactions on Intelligent Systems and Technology}, 14(1):1--22, 2022.

\bibitem[\protect\citeauthoryear{Wang \bgroup \em et al.\egroup }{2022b}]{ShuttleNet_AAAI_2022}
Wei{-}Yao Wang, Hong{-}Han Shuai, Kai{-}Shiang Chang, and Wen{-}Chih Peng.
\newblock Shuttlenet: Position-aware fusion of rally progress and player styles for stroke forecasting in badminton.
\newblock In {\em {AAAI}}, pages 4219--4227. {AAAI} Press, 2022.

\end{thebibliography}

\end{document}